\documentclass[letterpaper]{article} % DO NOT CHANGE THIS
\usepackage[]{aaai24_arxiv}  % DO NOT CHANGE THIS
\usepackage{times}  % DO NOT CHANGE THIS
\usepackage{helvet}  % DO NOT CHANGE THIS
\usepackage{courier}  % DO NOT CHANGE THIS
\usepackage[hyphens]{url}  % DO NOT CHANGE THIS
\usepackage{graphicx} % DO NOT CHANGE THIS
\usepackage{natbib}  % DO NOT CHANGE THIS AND DO NOT ADD ANY OPTIONS TO IT
\usepackage{caption} % DO NOT CHANGE THIS AND DO NOT ADD ANY OPTIONS TO IT
\usepackage{graphicx}
\usepackage{amsmath}
\usepackage{amssymb}
\newcommand{\etal}{\textit{et al.}}

\usepackage{algorithm}
\usepackage{algorithmic}

\usepackage{newfloat}
\usepackage{listings}
\DeclareCaptionStyle{ruled}{labelfont=normalfont,labelsep=colon,strut=off} % DO NOT CHANGE THIS
\floatstyle{ruled}
\newfloat{listing}{tb}{lst}{}
\floatname{listing}{Listing}
\title{Multiscale Residual Learning of Graph Convolutional \\ Sequence Chunks for Human Motion Prediction}
\author{
    Mohsen Zand \textsuperscript{\rm 1,2}, Ali Etemad \textsuperscript{\rm 1}, Michael Greenspan \textsuperscript{\rm 1}
}
\affiliations{
    \textsuperscript{\rm 1}Department of Electrical and Computer Engineering, and Ingenuity Labs Research Institute, \\
    Queen’s University, Kingston, ON, Canada.

    \textsuperscript{\rm 2} Research Computing Center, University of Chicago, \\ Chicago, Illinois, United States of America
}

\usepackage{bibentry}
\begin{document}

\maketitle

\begin{abstract}
A new method is proposed for human motion prediction by learning temporal and spatial dependencies.
Recently, multiscale graphs have been  developed to model the human body at higher  abstraction levels, resulting in more stable motion prediction. 
Current methods however predetermine scale levels and combine spatially proximal joints to generate coarser scales based on human priors, even though movement patterns in different motion sequences vary and do not fully comply with a fixed graph of spatially connected joints. Another problem with graph convolutional methods is mode collapse, in which predicted poses converge around a mean pose with no discernible movements, particularly in long-term predictions. To tackle these issues, we propose \emph{ResChunk}, an end-to-end network which explores dynamically correlated body components based on the pairwise relationships between all joints in individual sequences. ResChunk  is trained to learn the residuals between target sequence chunks in an autoregressive manner to enforce the temporal connectivities between consecutive chunks. 
It is hence a sequence-to-sequence prediction network which considers dynamic spatio-temporal features of sequences at multiple levels. 
Our experiments on two challenging benchmark datasets, CMU Mocap and Human3.6M, demonstrate that our proposed method is able to effectively model the sequence information for motion prediction and outperform other techniques to set a new state-of-the-art. Our code is available at \url{https://github.com/MohsenZand/ResChunk}.
\end{abstract}

%%%%%%%%% BODY TEXT
\section{Introduction}
\label{sec:introduction}

%Prediction of short-term future motion is easily achieved by humans, such as when driving or navigating through crowded scenes, but remains a challenging task to automate due to the inherent dynamic and stochastic nature, non-linearity, high dimensionality, and complex context-dependency of human motion. Despite these inherent challenges, human motion prediction has been an active area of research in computer vision due to its useful applications in autonomous driving, human-robot interaction, healthcare, surveillance, and many other applications~\cite{gui2018adversarial,martinez2017human,liu2019towards}. 

The human motion prediction problem aims to estimate a graph of human joint values at some future time. When modeling human motion, the positions of the joints in the predicted frames generally depend on both the temporal smoothness and the bio-mechanical dynamics of human motion  ~\cite{gui2018adversarial,li2021multiscale,dang2021msr,zand2023flow}. 
Prior studies used different strategies, including Recurrent Neural Networks (RNNs)~\cite{fragkiadaki2015recurrent,guo2019human,martinez2017human}, Generative Adversarial Networks (GANs)~\cite{kundu2019bihmp,gui2018adversarial,cui2021efficient}, and Graph Convolutional Networks (GCNs)~\cite{mao2019learning,li2020dynamic}. In particular, the use of GCNs with learnable adjacency matrices~\cite{mao2019learning,li2020dynamic,cui2020learning,mao2020history,dang2021msr} to encode the spatial structure of human pose has yielded encouraging results.  

\begin{figure}[t]
    \centering
    \includegraphics[width=1.0\linewidth]{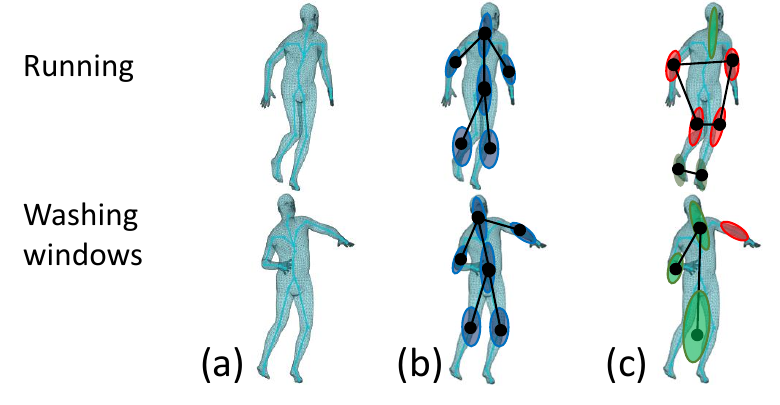}
    \caption{(a) Human Joints.
    %Original joint scale. 
    (b) Manually designated grouping of spatially close joints in DMGNN~\cite{li2020dynamic}, MSR-GCN~\cite{dang2021msr}, and MST-GNN~\cite{li2021multiscale}. (c) Automatically grouped correlated joints in the proposed method, with each color representing one group. Even distant joints not directly connected by the kinematic tree may be automatically grouped together if they similarly participate in a particular movement. 
    % Each color corresponds to one group in our proposed method. 
    }
    \label{fig:multiscale}
\end{figure}

More recent methods try to model increasingly comprehensive relations by representing the human body at multiple scales~\cite{li2020dynamic,li2021multiscale,dang2021msr}. They group spatially proximal joints together and represent each group by a coarser pose. The motion patterns of the body and movement coordination are thus better stabilized. The coarser level motion is demonstrated to be more stable, allowing for easier pose prediction. 
These methods however initialize multiscale body graphs by merging spatially proximal joints to create coarser scales based on manually designated priors. 
This is illustrated in Figure~\ref{fig:multiscale} (b),  where  joints in close proximity are grouped together and replaced with a single node. Apparently, movement patterns do not fully comply with the spatial connectivity of joints and rather distant, less proximal joints may exhibit a greater correlation for some movements. 
The collaborative motions of arms and legs, for example, are 
% easily comprehended 
more highly correlated 
in the action of `running'
than some 
more proximal joints,
such as wrists to arms and feet to legs.
% joints and feet joints (to legs), 
Thus, stable motions necessitate synchronizing diverse body components, including those that are not adjacent, or closely connected by the kinematic tree~\cite{li2018convolutional,mao2019learning}.

Another common issue is over-smoothing, 
also known as mode collapse,
where predicted poses tend to converge to a single mean pose with no discernible motions. This generally occurs in deep networks, where joint features are averaged to increase similarity with the ground-truth during information propagation. 

To tackle these issues, we propose \emph{ResChunk}, a novel human motion prediction method which is trained to learn 
residuals,
which form skip connections 
between consecutive target sequence chunks as well as dynamic action-agnostic multiscale representations for individual sequences. 
Specifically, our method is a neural graph solution which can directly model the log-likelihood of temporal and spatial information for relatively long sequences. 
%We evaluate our method by performing extensive experiments and ablation studies on CMU~\cite{de2009guide} and H3.6M~\cite{ionescu2013human3} datasets, and illustrate that ResChunk outperforms other methods in the field by generating more accurate and consistent predictions both qualitatively and quantitatively.
Our contributions are summarized as follows:
\begin{itemize}
    \item We propose a trainable algorithm for automatic grouping of correlated joints. It models the human body using a fully connected graph neural network and explores the edge strengths by learning pairwise interactions among all joints. The learned interactions constitute a correlation matrix, which we use to group the most correlated joints together as body components. 
    
    \item We propose a novel algorithm for sequence-to-sequence prediction, which learns the residuals between consecutive target sequence chunks. 
    The method allows for temporal smoothness while avoiding the over-smoothing issue. It specifically enforces the spatio-temporal connectivities in an efficient autoregressive manner, eliminating the sequential motion state computations. 
    
    \item We apply our method to the challenging task of motion prediction and demonstrate that our model is able to leverage both intra-frame and inter-frame dependencies between joints and correlated groups of joints across frames. We perform extensive experimental evaluations on two widely used challenging datasets and show the effectiveness of the proposed method for skeleton-based motion prediction by setting a new state-of-the-art. 
    
    %\item We also make our implementation public
    %~\footnote{https://github.com/MohsenZand/ResChunk} 
    %to enable reproducibility and contribute to the field. 
\end{itemize}

We also release our code at:\\ \url{https://github.com/MohsenZand/ResChunk}
%We will also make our source code public upon acceptance of this paper.

%%%%%%%%%%%%%%%%%%%%%%%%%%%%%%%%%%%%%%%%%%%
\section{Related Work}
\label{sec:related_work}

%In several previous methods, sequential joint features were modeled using different variations of recurrent neural networks (RNNs) given their ability to learn time series~\cite{wang2012mining,fernando2015modeling}. For instance, Martinez~\etal~\cite{martinez2017human}  developed a sequence-to-sequence architecture with residual connections by applying changes such as sample-based loss to the standard RNN. Comprehensive motion modeling, however, requires both spatial and temporal reasoning, while RNN-based methods are generally unable to model the spatial relationships between the joints. Moreover, the training of an RNN model easily collapses via gradient explosion. Discontinuities are also common due to the frame-wise prediction style.

%Generative Adversarial Networks (GANs)~\cite{kundu2019bihmp,gui2018adversarial,cui2021efficient} can create realistic data with comparable patterns to the training data. They are however vulnerable, unstable for larger outputs, and difficult to train~\cite{pathak2016context}. Transformer-based methods~\cite{aksan2021spatio,cai2020learning} demonstrate the inherent advantage of modeling intra- and inter-joint dependencies over RNN, but they are computationally expensive.  

Leveraging the structure and spatial relationships between joints has recently shown encouraging results, suggesting the significance of the connectivity~\cite{
%aksan2019structured,
singh2020multi,lu2020structured}.
%In~\cite{aksan2019structured}, a structured prediction layer was introduced to model the joint dependencies given motion context. Specifically, a hierarchy of small neural networks was used for individual joints, similar to the kinematic chains of the human skeleton.
%Cai~\etal~\cite{cai2020learning} proposed to model structural configurations by progressively predicting the joint trajectories based on the kinematic connections. They also leveraged a transformer network to model the spatial and temporal dependencies of human motions. 
Mao~\etal~\cite{mao2019learning} modeled spatial structure using graph convolutional networks (GCNs) and encoded temporal information in trajectory space via the Discrete Cosine Transform (DCT). They adopted the DCT to smooth the 3D coordinates of human motion by discarding high frequencies. 
The DCT was employed in several other approaches~\cite{li2021skeleton,mao2020history} to transfer body-joint poses across time to the frequency domain. 

To capture a functional group of joints or high-order relationships, multiscale modeling was proposed. For instance, Du~\etal~\cite{du2015hierarchical}, split the human skeleton into five body parts based on the physical structure of the human body. These were then fed into five bidirectional RNNs to automatically learn a multi-level representation of skeletal data. In~\cite{zhu2016co}, an RNN model using a deep Long Short-Term Memory (LSTM) network was developed for skeleton-based action recognition. Their deep LSTM network modeled co-occurrences between joints. 
%The approach of Shahroudy~\etal~\cite{shahroudy2016ntu} considered human activities as interactions of different body parts and developed a part-aware LSTM model. They utilized long-term temporal correlations for each body part and split the memory cell of the LSTM into part-based sub-cells. 

Dang~\etal~\cite{dang2021msr} recently proposed a multiscale residual graph convolution network (MSR-GCN) that cast the human pose as a fully connected graph, and used graph convolution networks to learn the relations between all pairs of joints. They also abstracted a human pose by grouping neighboring joints together to obtain more stable motions in coarser poses.  Li~\etal~\cite{li2021multiscale} also proposed to embed underlying features at individual scales and then fused features across scales to obtain a comprehensive representation.

These multiscale architectures however predetermine scale levels and combine spatially
close joints to generate coarser scales based on manually designated priors. Apparently, diverse motion sequences have different movement patterns that do not entirely conform to a defined graph of spatially adjacent joints. These methods are also unable to model the spatio-temporal correlations over multiple frames. 
They typically use hand-crafted rules to incorporate spatial characteristics and RNNs to capture temporal dependencies.
%Hence, learning the spatial dependencies is decoupled from the temporal relationships.

Similar to our approach, Cui~\etal~\cite{cui2020learning} parameterized the adjacent matrix as the connective graph, and learned the weights of natural connections instead of fixed ones. They proposed a learnable global graph to capture implicit relationships. Another multiscale architecture was proposed by Li~\etal~\cite{li2020dynamic} to model the internal relationships among body joints. They introduced a multiscale graph
computational unit, which leveraged a multiscale graph to extract and fuse features across multiple scales. They also proposed a graph-based GRU to enhance state propagation for pose generation. 

In contrast, our method is trained to predict dynamic structures in individual sequences and simultaneously considers temporal dependencies among sequence chunks via learning residuals,
which we show experimentally to improve accuracy.

%%%%%%%%%%%%%%%%%%%%%%%%%%%%%%%%%%
\section{Proposed Method}
\label{sec:proposed_method}

\subsection{Problem Formulation}
Let $X_{1:S}\!=\!\{X_1, \dots, X_S\}$ denote a motion sequence where each frame $X_t\!\in\!\mathbb{R}^K$, $K\!=\!J\times \!D$, represents a human body pose at time $t$, with  $J$ joints in the skeleton, and each joint comprising $D$ dimensions. Each joint can be represented by $D\!\!=\!\!3$ indicating a pure 3D translation, or can more generally include a rotational component indicating the orientation of a coordinate reference frame centered at each joint, represented as a matrix ($D\!\!=\!\!9$), angle-axis ($D\!\!=\!\!3$), or quaternion ($D\!\!=\!\!4$). Thus, each joint $j$ at time-step $t$ can be represented by $X^j_t \!\!\in\!\! \mathbb{R}^D$,
i.e.
$X_t\! = \!\{X^1_t, X^2_t \ldots  X^J_t\}$.

In human motion prediction, the goal is to predict the future pose sequence $y\!=\!\{X_{T+1:T+p}\}$ consisting of $p$ frames given the observed sequence $x\!=\!\{X_{1:T}\}$. We thus aim to model a mapping from input poses at time-steps $i=1\dots T$ to a structured output consisting of poses at frames $i=(T\!+\!1),
(T\!+\!2) \ldots (T\!+\!p)$. 

\begin{figure*}[!t]
    \centering
    \includegraphics[width=0.9\linewidth]{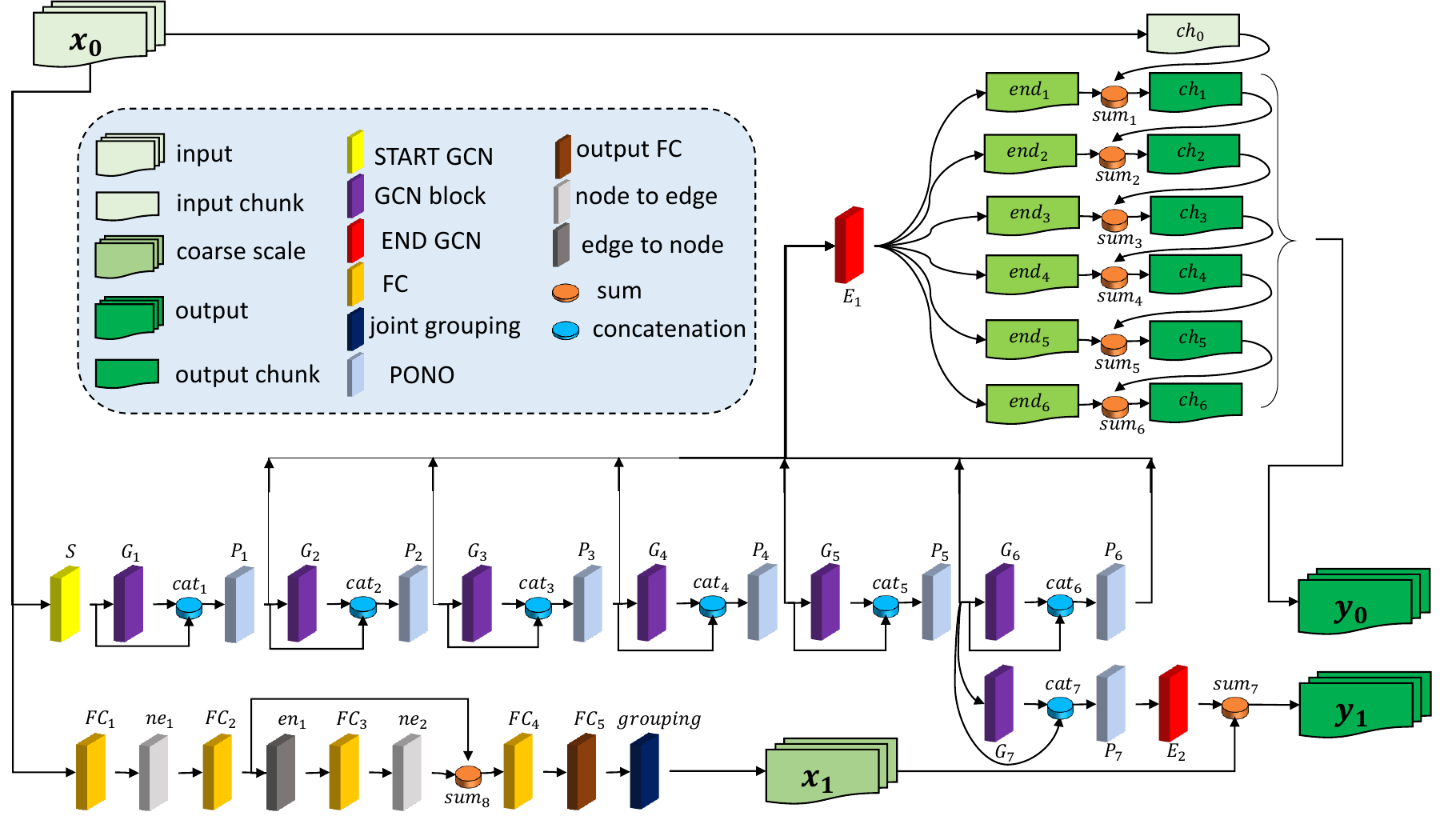}
    \caption{Schematic overview of the proposed architecture for multiscale motion prediction.
    The model is trained to learn residuals between output sequence chunks, automatic grouping of the correlated body joints, and residuals between input and output sequences in the coarse scale. The output sequence $y_0$ is divided into multiple arbitrary sized chunks, and the pose displacement at each chunk is predicted by a residual connection from the previous chunk at the final block. 
    We automatically generate the coarse scale $x_1$ from correlated body components using the input motion sequence $x_0$ in the joint level scale. Our joint grouping strategy is trainable and flexible for individual sequences. }
    \label{fig:model}
\end{figure*}

We model the human body using two graphs at different scales. 
The first graph encodes the human skeleton with the original joint values as nodes, while the second graph represents rich body interactions among body components by modeling groups of joints as individual nodes. While the first graph is naturally built from the input motion sequence, the correlated joints are learned and grouped together to construct the nodes in the second graph. The trainable joint grouping and multiscale motion prediction are both achieved in an end-to-end network, illustrated in Figure~\ref{fig:model}. 
The motion sequence in the original scale
%, \it{i.e.}, 
$x_0$, is used to learn the sequence $x_1$ in the coarser scale. These two sequences are used as input observed motions to predict the future motion sequences $y_0$ and $y_1$. In the following subsections, we describe the details of our proposed method.

%-----------------------------------------
\subsection{Residual Learning of Sequence
Chunks}

The network input is $x\!=\!\{X_{1:T}\}$, which is considered as the joint level scale. We denote it as $x_0$ to distinguish it from the sequence in the coarse scale $x_1$. Hence, $x_0$ consists of trajectories of the feature vectors for $J$ nodes in the observed frames $1\dots T$. The input dimension is thus $T \times K$, where $K\!=\!J\! \times \! D$. 

We treat human pose as a fully connected graph and use the capacity of Graph Convolutional Networks (GCNs)~\cite{kipf2016semi} to dynamically and flexibly learn the relationships between all pairs of joints. The GCNs usually include multiple stacked graph convolutional layers. Each layer $l$ takes a matrix $H^l \in \mathbb{R}^{K\times F^l}$ as input, with $F$ denoting the number of features
output by the previous layer, and returns a matrix of the following form:
\begin{equation}
H^{l+1}=\sigma(A^lH^lW^l)
\end{equation}
where $H^{l+1} \in \mathbb{R}^{K\times F^{l+1}}$, $A^l\in \mathbb{R}^{K\times K}$ denotes the weighted adjacency matrix, $W^l \in \mathbb{R}^{F^l \times F^{l+1}}$ denotes the trainable parameters, and $\sigma(.)$ is an activation function, such as $tanh(·)$. 
%Since all operations are differentiable, w.r.t. both $A^l$ and $W^l$, the resulting network can be trained using standard backpropagation.

In our method, the first GCN, denoted by $S$, takes the input $x_0 \in \mathbb{R}^{K \times T}$ and uses one graph convolutional layer to project the input pose to an intermediate representation of size $\mathbb{R}^{K \times F}$.
We then utilize $G_1$, a GCN block with 6 graph convolutional layers, each with the same input and output sizes as $\mathbb{R}^{K \times F}$. 
The input and output of the GCN block $G_1$ are concatenated and create an embedding of size $\mathbb{R}^{(2 \times K) \times F}$ which is employed in $P_1$, a PONO (position normalization)~\cite{li2019positional} block. 
PONO has been shown to extract position dependent statistics and reveal structural information. It is conducted based on positional statistics of mean and standard deviation to normalize the embeddings across the pose dimension. The following operations are performed in the PONO block:
\begin{equation}
\begin{aligned}
    &a, b = \textrm{split}(concat_1) \\
    &a = a - (mean(a) / std(a)) \\
    &pono_1 = a * \textrm{sigmoid}(b) \\
\end{aligned}
\end{equation}
where $concat_1$ is the output of the first concatenation module $cat_1$, and $\textrm{split}()$ partitions a tensor into two equally sized chunks $a$ and $b$ along the pose dimension. The output of the PONO block $P_1$ is used in three different blocks, $E_1$, $G_2$, and $cat_2$. The block $E_1$ employs one graph convolutional layer and outputs a tensor of size $\mathbb{R}^{K \times T}$, from which we separate $end_1$, a chunk of size $\mathbb{R}^{K \times c_1}$, where $c_1 = T / 6$. 
We then construct $ch_1$, the first chunk of the output sequence $y_0$, by using a residual connection between $end_1$ and $ch_0$ as:
\begin{equation}
    ch_1 = end_1 + ch_0
\end{equation}
where $ch_0$ is created from the last $c_1$ frames of the input motion sequence $x_0$. Specifically, $ch_0\!\!=\!\!\{X_{(T - c_1):T}\}$ and $ch_1\!=\!\{X_{(T +1):(T + c_1)}\}$. The same procedure is repeated to generate $\{ch_i\}_{i=2}^6$ as $ch_i\! = \!end_i \!+\! ch_{i-1}$. The output %sequence 
$y_0$ is then constructed by integrating $ch_1,\dots, ch_6$. Note that all intermediate GCN blocks and PONO modules utilize the same architecture as described for $G_1$ and $P_1$, respectively. 

The extracted features through $S$ to $E_1$ are decoded to poses at the joint scale level. The proposed architecture learns the residual vectors between the corresponding sequence chunks in an efficient autoregressive fashion in a single forward pass. 
The network is trained to learn the residuals between two successive chunks that are linked by a skip connection. It is obvious that adding a residual change to the current pose yields the future pose. We thus add residual connections between consecutive chunks to enforce the temporal connectivities. We avoid the sequential motion state computations of the recurrent-network-based models~\cite{martinez2017human,jain2016structural,fragkiadaki2015recurrent}, and directly predict the whole sequences. 
 
There exist other residual networks for motion prediction tasks which are fundamentally different from ours. 
For example, residual connections in~\cite{martinez2017human} are between the input and the output of each RNN cell, and both~\cite{mao2019learning} and~\cite{dang2021msr} add global skip connections between the input and the output poses of their networks.

To capture rich multiscale relationships and extract dynamic semantics from motion sequences, we further propose an automatic joint grouping method in the same end-to-end network, which can create highly flexible and trainable joint abstractions. In the following subsection, we detail our proposed approach.

%-----------------------------------------
\subsection{Trainable Human Joint Grouping}
Correlated body components are automatically extracted based on pairwise interactions among all human joints. In particular, we formulate the interactions between joints $i$ and $j$ as a latent variable $z_{ij} \in [0,1]$. The learned weights for all $z_{ij}$ constitute a correlation matrix, which we use to group the most correlated joints together as body components. 

In our network, we produce a factorized categorical distribution~\cite{kipf2018neural,graber2020dynamic,gong2021memory} as:
\begin{equation}
    q_\phi(z|x_0)=\Pi_{i \neq j} q_\phi (z_{ij} |x_0)
\end{equation}
where $\phi$ denotes the set of model parameters. Hence, the latent variable $z$ is conditioned on the input sequence $x$. The underlying graph on the joint interactions is however unknown. We thus use a fully connected graph neural network (GNN) %architecture 
to predict the latent graph structure. The model learns the relation probabilities for each edge connecting two arbitrary joints $i$ and $j$, which we later use to build a correlation matrix and then group the most correlated joints.

To construct the embedding 
$q_\phi(z_{ij}|x_0)$
for each edge, the input sequence $x_0$ is fed into a GNN that works on a fully connected graph with no self-loops. As shown in~\cite{kipf2018neural,graber2020dynamic}, the message passing operations on this graph can then be formulated as:
\begin{equation}
\begin{aligned}
    &h^{(1)}_j = f^{(0)}(r_j) \\
    v \rightarrow e: &h^{(1)}_{ij}=f^{(1)}_e([h^{(1)}_i,h^{(1)}_j])\\
    e \rightarrow v: &h^{(2)}_j=f^{(1)}_v(\sum_{i\neq j}h^{(1)}_{ij})\\
    v \rightarrow e: &h^{(2)}_{ij}=f^{(2)}_e([h^{(2)}_i,h^{(2)}_j])\\
\end{aligned}
\label{eq:messages}
\end{equation}
where $r_j\!=\!\{X_1^j, \dots, X^j_T\}$ denotes the trajectory of the $j$-th joint across the observed sequence; $v \!\in \!\mathcal{V}$, $e \!\in\! \mathcal{E}$, $h^{(l)}_i$ denotes the embedding of the node $v_i$ in the layer $l$, and; $h^{(l)}_{ij}$ represents the embedding of the edge $e_{(i,j)}$ in the layer $l$. The embedding $h^{(1)}_{ij}$ only relies on the trajectories of the joints $i$ and $j$, {\it ie.}, $r_i$ and $r_j$, disregarding interactions with other joints, whereas $h^{(2)}_j$ takes the information from the entire graph. The embedding $h^{(2)}_{ij}$ is finally used to obtain the edge connecting $v_i$ and $v_j$ as $q_\phi (z_{ij}|x_0)=\textrm{softmax}(h^{(2)}_{ij})$.  The embedding $h_{ij}^{(2)}$ is therefore the predicted logits for $z_{ij}$.

The functions $f^{()}_v$ and $f^{()}_e$ represent node- and edge- specific neural nets, which we model using multilayer perceptron (MLP) networks. 

In Eq.\ref{eq:messages}, we enable mapping from edge to node representations or vice versa through numerous rounds of message passing operations to model edge strength posterior as $q_\phi(z_{ij}|x_0)$. We then use a sampling procedure to generate the latent variable $z_{ij}$. 
The standard sampling from a categorical distribution is however nondifferentiable~\cite{graber2020dynamic}. Following~\cite{jang2016categorical,maddison2017concrete,kipf2018neural}, we instead use the concrete distribution, which is a continuous approximation of the discrete categorical distribution. Sampling is hence performed via reparameterization, which involves first sampling a vector $g$ from the $\textrm{Gumbel}(0, 1)$ distribution and then calculating $z_{ij}$ as:
\begin{equation}
    z_{ij} = \textrm{softmax}((h_{ij}^{(2)} + g)/\tau)
\label{eq:sampling}
\end{equation}
where $\tau$ stands for a temperature parameter that affects the distribution's smoothness. When $\tau \rightarrow 0$, the Gumble distribution converges to one-hot samples from the categorical distribution. 
As a result of this sampling technique, the model becomes differentiable, and the model weights can 
thus be updated using backpropagation.

From the sampled latent variables, we obtain the correlation matrix between all the joints. We then use a popular hierarchical agglomerative method~\cite{rokach2005clustering} to repeatedly
group the most correlated joints together. Once groups are identified, we use the average values of all joints in each group as their new pose value. We therefore do not reduce the number of joints in the groups at the coarse scale, and instead we use the same value for all group members. This procedure is performed for both input and output sequences during training. 

At inference, the network predicts the group structure and the averaged value for the joints in each group. 
Notably, the number of joint groups and the joints within each group may vary, since they are learned based on the movement patterns in each %motion 
sequence.
The proposed joint grouping method can create highly flexible and trainable joint abstractions, which we further use to capture rich multiscale relationships and extract dynamic semantics from motion sequences.

As noted above, we produce a factorized categorical distribution to automatically generate the coarse scale $x_1$ from correlated body components using the input motion sequence $x_0$. 
The coarse scale is hence a latent representation which is learned directly from the data, and so there is no ground truth for scale. It is specifically obtained by clustering the  correlation matrix, which is the relation probabilities for each edge connecting two arbitrary joints $i$ and $j$. These probabilities are basically latent interactions between the joints in the latent graph structure. Since the joint interactions in the latent graph are unknown, there is no ground-truth for the coarse scale.

\subsection{Training and Inference}

Given a training sequence $x_0$, we compute $q_\phi(z_{ij}|x_0)$ and sample $z_{ij}$ using Eq.\ref{eq:sampling}. The KL-divergence $\textrm{KL}[q_\phi(z|x_0)\|p_\theta(z)]$ between the uniform prior and the predicted approximate posterior is calculated as:
\begin{equation}
    \sum_{i \neq j} H(q_\phi(z_{ij}|x)) + c
\end{equation}
where $H$ denotes the entropy function, and $c$ is a constant since the prior $p_\theta(z)\!=\!\Pi_{i \neq j} p_\theta(z_{ij})$ is a factorized uniform distribution over edges types. 

The predicted outputs $\hat{y}_0$ and $\hat{y}_1$ are assumed to be the means of a Gaussian distribution with a fixed variance. The overall training objective is to minimize negative ELBO as:
\begin{equation}
\textstyle
%\begin{aligned}
    \textrm{KL}[q_\phi(z|x_0)\|p_\theta(z)]-(\sum_i \frac{\|x_{0_i} - \hat{x}_{0_i}\|}{2\sigma_0^2} + \\
    \sum_i \frac{\|x_{1_i} - \hat{x}_{1_i}\|}{2\sigma_1^2}+ c)
%\end{aligned}
\end{equation}
where $c$ is a constant. 

At inference time, we aim to predict $\hat{y}_0$ and $\hat{y}_1$ given the input sequence $x_0$. We use the prior to model latent variable $z$ and predict the relation types between the joints. 
We then calculate the correlation matrix and group correlated joints together. This is accomplished by assigning the average pose value of the joints in each group to all the joints in that group.

%-----------------------------------------------------
\section{Experiments}
\label{sec:experiments}
%We evaluate the quantitative performance of the proposed method on two challenging benchmark datasets, CMU motion capture (CMU Mocap available at http://mocap.cs.cmu.edu/),
%~\footnote{http://mocap.cs.cmu.edu/}, 
%and Human3.6M (H3.6M)~\cite{ionescu2013human3}. 
%, and archive of motion capture as surface shapes (AMASS) ~\cite{mahmood2019amass}. 
%We provide comparisons to previous state-of-the-art motion prediction techniques. We also present ablation experiments on the key components of our model to quantify their impact. 

%==============================================
\begin{table*}
\centering \scriptsize
\setlength{\tabcolsep}{2.5pt}
\begin{tabular}{l|ccccc|ccccc|ccccc|ccccc}
\hline
actions & \multicolumn{5}{|c|}{basketball} & \multicolumn{5}{|c|}{basketball signal} & \multicolumn{5}{|c|}{directing traffic} & \multicolumn{5}{|c}{jumping}
 \\
\hline
milliseconds & 80 & 160 & 320 & 
400 & 1000 & 80 & 160 & 320 & 
400 & 1000 & 80 & 160 & 320 & 
400 & 1000 & 80 & 160 & 320 & 
400 & 1000 \\
\hline
 
Seq2seq%~\cite{martinez2017human}
& 15.45 & 26.88 & 43.51 & 
49.23 & \textbf{72.83} & 20.17 & 32.98 & 42.75 & 
44.65 & 60.57 & 20.52 & 40.58 & 75.38 & 
90.36 & 153.12 & 26.85 & 48.07 & 93.50 & 
108.90 & 162.84 \\

TrajDep%~\cite{mao2019learning}
& 11.68 & 21.26 & 40.99 & 
50.78 & 97.99 & 03.33 & 06.25 & 13.58 & 
17.98 & 54.00 & \underline{06.92} & 13.69 & 30.30 & 
39.97 & 114.16 & 17.18 & 32.37 & 60.12 & 
72.55 & 127.41 \\

DMGNN%~\cite{li2020dynamic} 
& 15.57 & 28.72 & 59.01 & 
73.05 & 138.62 & 05.03 & 09.28 & 20.21 & 
26.23 & 52.04 & 10.21 & 20.90 & 41.55 & 
52.28 & 111.23 & 31.97 & 54.32 & 96.66 & 
119.92 & 224.63 \\

MSR-GCN%~\cite{dang2021msr}
& \underline{10.28} & \underline{18.94} & \underline{37.68} & 
\underline{47.03} & {86.96} & 03.03 & 05.68 & 12.35 & 
16.26 & \underline{47.91} & 05.92 & 12.09 & 28.36 &
38.04 & \underline{111.04} & 14.99 & \underline{28.66} & \underline{55.86} & 
\underline{69.05} & \underline{124.79} \\

SGSN%~\cite{li2021skeleton}
& 11.10 & 20.20 & 41.30 & 
52.90 & 89.10 & \textbf{02.20} & \textbf{04.10} & \underline{09.70} & 
\textbf{14.70} & 51.50 & 05.80 & \textbf{11.00} & \underline{24.70} & 
\textbf{32.10} & 137.60 & \underline{13.80} & 29.90 & 071.50 & 
90.80 & 160.20 \\

Ours (ResChunk) 
& \textbf{10.18} & \textbf{17.42} & \textbf{37.33} & 
\textbf{45.28} & \underline{86.88} & \underline{02.85} & \underline{05.54} & \textbf{09.27} & 
\underline{15.55} & \textbf{45.23} & \textbf{05.82} & \underline{11.71} & \textbf{24.10} & 
\underline{33.97} & \textbf{110.35} & \textbf{13.64} & \textbf{25.07} & \textbf{52.03} & 
\textbf{63.37} & \textbf{121.38} \\\hline\hline

actions & \multicolumn{5}{|c|}{running} & \multicolumn{5}{|c|}{soccer} & \multicolumn{5}{|c|}{walking} & \multicolumn{5}{|c}{washing window}
 \\
\hline

milliseconds & 80 & 160 & 320 & 
400 & 1000 & 80 & 160 & 320 & 
400 & 1000 & 80 & 160 & 320 & 
400 & 1000 & 80 & 160 & 320 & 
400 & 1000 \\
\hline

Seq2seq%~\cite{martinez2017human}
& 25.76 & 48.91 & 88.19 & 
100.80 & 158.19 & 17.75 & 31.30 & 52.55 & 
61.40 & 107.37 & 44.35 & 76.66 & 126.83 & 
151.43 & 194.33 & 22.84 & 44.71 & 86.78 & 
104.68 & 202.73 \\

TrajDep%~\cite{mao2019learning} 
& 14.53 & 24.20 & 37.44 & 
41.10 & 51.73 & 13.33 & 24.00 & 43.77 & 
53.20 & 108.26 & 06.62 & 10.74 & \underline{17.40} & 
20.35 & 34.41 & 05.96 & 11.62 & \underline{24.77} &
\underline{31.63} & \textbf{66.95} \\

DMGNN%~\cite{li2020dynamic} 
& 17.42 & 26.82 & 38.27 & 
40.08 & 46.40 & 14.86 & 25.29 & 52.21 & 
65.42 & 111.90 & 09.57 & 15.53 & 26.03 & 
30.37 & 67.01 & 07.93 & 14.68 & 33.34 & 
44.24 & 82.84 \\

MSR-GCN%~\cite{dang2021msr}
& \underline{12.84} & \underline{20.42} & 30.58 & 
34.42 & 48.03 & 10.92 & 19.50 & \underline{37.05} & 
\underline{46.38} & \underline{99.32} & 06.31 & 10.30 & 17.64 & 
\underline{21.12} & 39.70 & 05.49 & 11.07 & 25.05 &
32.51 & 71.30 \\

SGSN%~\cite{li2021skeleton}
& 19.80 & 24.70 & \textbf{26.60} & 
\underline{30.20} & \textbf{44.20} & \textbf{09.20} & \underline{18.10} & 38.80 & 
49.50 & 103.30 & \underline{05.90} & \textbf{09.60} & 18.60 & 
22.80 & \textbf{31.20} & \textbf{04.90} & \underline{10.10} & 28.10 & 
37.30 & 71.10 \\

Ours (ResChunk) 
& \textbf{12.35} & \textbf{18.56} & \underline{28.56} & 
\textbf{30.17} & \underline{46.25} & \underline{10.28} & \textbf{17.65} & \textbf{35.56} & 
\textbf{45.10} & \textbf{98.62} & \textbf{05.74} & \underline{09.84} & \textbf{17.27} & 
\textbf{20.84} & \underline{33.90} & \underline{05.24} & \textbf{10.03} & \textbf{24.25} & 
\textbf{30.45} & \underline{70.35} \\\hline
\end{tabular}

\caption{Prediction results over 8 action categories of the CMU Mocap dataset. The results are reported in terms of MPJPE (smaller is better). The best performance is in boldface, and the second-best performance is underlined. }
\label{table:cmu}
\end{table*}
%==============================================

%-----------------------------------------------------
\subsection{Experiment Setup}
%In this subsection, we describe the benchmark datasets, evaluation metrics, and implementation details of ResChunk. 

%-----------------------------------------------------
\subsubsection{Datasets and Protocols}
\emph{CMU Mocap}~\cite{de2009guide}~\footnote{http://mocap.cs.cmu.edu/} is a challenging dataset containing 8 action categories performed by 144 subjects. 
Each pose is represented by 38 body joints, among which we utilize 24 joints. 
%in our experiments.

\emph{H3.6M}~\cite{ionescu2013human3} is a large dataset containing 3.6 million high quality 3D body joint positions. The video sequences are captured by a Vicon motion capture system and sampled at 50 frames per second (fps). 
At each frame, the 3D position of 32 body joints are captured, from which we choose
a subset of 21 representative joints.
%, similar to ~\cite{jain2016structural, martinez2017human}, and~\cite{aksan2019structured}. 
This dataset is collected for 15 periodic and non-periodic actions performed by seven different professional actors. The actors are named as S1, S5, S6, S7, S8, S9, and S11.
Following the literature~\cite{aksan2019structured,%dang2021msr,
sofianos2021space}, we use the data of S5 and S11 as test and validation datasets, respectively. The rest is used for training.

%\noindent\textbf{AMASS}~\cite{mahmood2019amass} is a more recent dataset consisting of multiple smaller publicly available motion datasets, \eg  HumanEva~\cite{sigal2010humaneva} and CMU Mocap~\cite{de2009guide}. It is a challenging dataset which provides many more (14$\times$) samples than H3.6M captured for a diverse range of actions. It uses a skinned multi-person linear (SMPL)~\cite{loper2015smpl} model for motion representation. The dataset includes $8,593$ clips consisting of $9,084,918$ frames sampled at 60 fps. We use 90\%, 5\%, and 5\% of all the clips for training, validation, and testing, respectively. We use 15 major joints for motion representation, as in~\cite{aksan2019structured}. 

We use an angle-axis representation with $D=3$ in both datasets. We also utilize a sliding window strategy over the sequences. In particular, we extract sequences with the length of 3 seconds 
%(180 frames if FPS = 60, else 
(150 frames if FPS = 50), and the stride is selected as 
%30 frames in AMASS dataset, and 
10 frames in both H3.6M and CMU Mocap datasets. Our data generator then creates a 2-second window initiated from a random frame in the 3-second sequence. The first second is used as the input sequence, and we predict the following 1-second sequence.

%-----------------------------------------------------
\subsubsection{Evaluation Metric}
The commonly used metric MPJPE (Mean Per Joint Position Error)~\cite{dang2021msr,
%li2020dynamic,
mao2019learning} is used to evaluate the performance of all techniques. It calculates the Euclidean distance between the 3D position of each estimated joint compared to the corresponding target position. We report this metric in millimeters for all experiments on both datasets. 
% We calculate the mean values by summing up $r$ time-steps for each joint. 
We report $r\!=\!80, 160, 320, 400$ milliseconds for short-term prediction, and $r=1000$ milliseconds for long-term prediction.

%-----------------------------------------------------
\subsubsection{Implementation Details}

%In Table~\ref{table:convs}, we show the detailed network parameters of the conditioning module in ResChunk architecture. 
We implement our models using PyTorch.
% and will make the code publicly available upon acceptance of this paper, to reproduce the results. 
All models are trained using Adam optimizer~\cite{kingma2014adam} with initial learning rate 0.0002, weight decay 0.0005, $\beta_1=0.9$, and $\beta_2=0.999$. The network is trained with a batch size of 128 on a single NVIDIA TITAN RTX GPU. We employ early stopping to the network for all the experiments. 
The hidden size in $FC_1$, $FC_2$, $FC_3$, $FC_4$, and $FC_5$ is set to 256, and the temperature parameter is $\tau=0.5$.

%-----------------------------------------------------
\subsection{Results}
In this section, we report the evaluation results on CMU Mocap and H3.6M datasets. We use 
%RNN (baseline), %zero-Velocity~\cite{martinez2017human},
seq2seq~\cite{martinez2017human}, %SRNN~\cite{jain2016structural}, 
TrajDep~\cite{mao2019learning} (trajectory dependencies),
DMGNN~\cite{li2020dynamic} (dynamic multiscale graph neural networks), %RNN-SPL~\cite{aksan2019structured},
MSR-GCN~\cite{dang2021msr} (multiscale residual graph convolution networks), and 
SGSN (skeleton graph scattering networks)~\cite{li2021skeleton}
%HRI~\cite{mao2020history},
%STSGCN~\cite{sofianos2021space}, and %DLOW~\cite{yuan2020dlow} 
for comparison,
as they are recent competitive %related 
methods.

%We use a single LSTM cell with 1024 units in the RNN model. We set a hidden output layer of size 960, and use rotation matrix representation for this model. 
For seq2seq model, we use residual supervised multi-action (Residual sup. (MA) in~\cite{martinez2017human}), where the decoder predictions are fed back to the model in a sampling-based mode. For the other methods, we report the results from their original papers.

%We run these methods on the same machine with their respective evaluation protocols for a fair comparison. 
% We use angle-axis and quaternion representations for seq2seq and QuaterNet, respectively. For the other method, rotation matrix representation was utilized.

%==============================================
\begin{table*}
\centering \scriptsize
\setlength{\tabcolsep}{2.5pt}
\begin{tabular}{l|ccccc|ccccc|ccccc|ccccc}
\hline
 & %\multicolumn{4}{|c|}{walking} & 
 \multicolumn{5}{|c|}{average on all 15 actions} &\multicolumn{5}{|c|}{eating} & %\multicolumn{5}{|c|}{smoking} &
 \multicolumn{5}{|c}{discussion}
 & \multicolumn{5}{|c}{directions}
 \\
\hline
milliseconds & 
80 & 160 & 320 & 
400 & 1000 &
80 & 160 & 320 & 
400 & 1000 & 80 & 160 & 320 & 
400 & 1000 & 80 & 160 & 320 & 
400 & 1000 \\
\hline

Seq2seq%~\cite{martinez2017human}
%& 29.36 & 50.82 & 76.03 & 81.51 
& 34.66 & 61.97 & 101.08 & 115.49 & 137.96
& 16.84 & 30.60 & 56.92 & 
68.65 & 100.20 & %22.96 & 42.64 & %70.14 & 
%82.68 & 94.83 & 137.44 & 
32.94 & 61.18 & 90.92 & 
96.19 & 161.70 & 35.36 & 57.27 & 76.30 & 
87.67 & 152.48 \\

TrajDep%~\cite{mao2019learning}
%& 12.29 & 23.03 & 39.77 & 46.12 
& 12.68 & 26.06 & 52.27 & 63.51 & 93.42
& 08.36 & \underline{16.90} & \underline{33.19} & 
40.70 & 77.75 &  %07.94 & 16.24 & %31.90 & 
%38.90 & 50.74 & 72.62 & 
12.50 & 27.40 & 58.51 & 
71.68 & 121.53 & \underline{08.97} & 19.87 & 43.35 &
\underline{53.74} & 101.79 \\

DMGNN%~\cite{li2020dynamic} 
%& 17.32 & 30.67 & 54.56 & 65.20 
& 16.95 & 33.62 & 65.90 & 79.65 & 103.22
& 10.96 & 21.39 & 36.18 & 
43.88 & 86.66 & %08.97 & 17.62 & %32.05 & 
%40.30 & 50.85 & 72.15 & 
17.33 & 34.78 & 61.03 & 
69.80 & 138.32 & 13.14 & 24.62 & 64.68 & 
81.86 & 115.75 \\

MSR-GCN%~\cite{dang2021msr}
& 12.11 & 25.56 & 51.64 & 62.93 & 91.73
%& 12.16 & 22.65 & 38.64 & 45.24 
& 08.39 & 17.05 & \textbf{33.03} & 
\underline{40.43} & 77.11 & %08.02 & 16.27 & %31.32 &
%38.15 & 49.45 & 71.64 & 
11.98 & 26.76 & 57.08 & 
69.74 & 117.59 & \textbf{08.61} & \textbf{19.65} & \underline{43.28} & 
53.82 & \underline{100.59} \\

SGSN%~\cite{li2021skeleton}
& \underline{10.65} & \underline{22.90} & \underline{47.00} & \underline{56.90} & \textbf{81.13}
%& \underline{08.30} & \underline{15.00} & \textbf{26.70} & \textbf{31.50} 
& \underline{07.90} & 17.40 & 35.80 & 
43.70 & \textbf{68.20} & %\textbf{07.00} & \underline{13.80} & %\textbf{23.60} & 
%\textbf{28.20} & \textbf{31.60} & \textbf{58.80} & 
\underline{08.00} & \underline{19.10} & \textbf{34.70} & 
\textbf{40.40} & \underline{96.50} & 10.60 & 21.60 & 45.40 &
56.30 & 101.00 \\

Ours (ResChunk) 
& \textbf{09.15} & \textbf{21.40} & \textbf{45.69} & \textbf{56.41} & \underline{81.95}
%& \textbf{08.25} & \textbf{14.52} & \underline{26.91} & \underline{40.17} 
& \textbf{07.62} & \textbf{15.38} & {33.50} & 
\textbf{40.08} & \underline{70.54} & %\underline{07.85} & \textbf{13.41} & %\underline{26.65} & 
%\underline{33.94} & \underline{42.93} & \underline{63.25} & 
\textbf{07.20} & \textbf{18.25} & \underline{37.82} & 
\underline{45.16} & \textbf{95.36} & {09.32} & \underline{19.71} & \textbf{42.86} & 
\textbf{51.76} & \textbf{98.65} \\  \hline

\end{tabular}
\caption{Short-term and long-term prediction results over 3 representative action categories of H3.6M dataset. The average prediction results across all 15 actions are also shown. The results are reported in terms of MPJPE (smaller is better). The best performance is in boldface, and the second-best performance is underlined.}
\label{table:h36m_short_long}
\end{table*}
%==============================================

%-----------------------------------------------------
\subsubsection{Evaluation on CMU Mocap}

We compare our model with state-of-the-art methods on all 8 activities, including \emph{basketball, basketball signal, directing traffic, jumping, running, soccer, walking} and \emph{washing window}. 

The quantitative comparisons in terms of MPJPE are shown in Table~\ref{table:cmu}. It can be seen that our method achieves the best or the second best results in all considered activities. On the `jumping',
%action, 
for instance, ResChunk outperforms existing methods for both short-term and long-term predictions. 
For $r\!\!=\!\!1000~ms$, it achieves 121.38 while the second best-performing method, 
% \ie, 
MSR-GCN, obtains 124.79. On the `running',
%action, 
as another example, our method yeilds 12.35 and 18.56 for $r\!=\!80$ and $r\!=\!160$, respectively, which are better than those of MSR-GCN with 12.84 and 20.42.

\subsubsection{Evaluation on H3.6M}

Table~\ref{table:h36m_short_long} reports the prediction results for 3 representative action categories of the H3.6M dataset. %As shown, 
Our method achieves the best performance in most time horizons for all activities. 
%On the `greeting' category, for example, ResChunk yields smaller MPJPE errors of 11.57, 25.70, 61.92, 77.94 respectively for $r\!\!=\!\!80$, $160$, $320$, and $400$, which outperforms SGSN~\cite{li2021skeleton} with 12.60, 26.50, 63.80, 79.60 respectively in the same time horizons. 
On average, ResChunk improves the short-term prediction significantly by obtaining the best performance 
for all $r\!\!=\!\!80, \!160, \!320$, and $400$ horizons as 9.15, 21.40, 45.69, 56.41, respectively, which outperforms SGSN %~\cite{li2021skeleton} 
with 10.65, 22.90, 47.00, 56.90 respectively in the same time horizons. Our method also yields 81.95 for $r\!\!=\!\!1000$, which is nearly as good as the top performance of 81.13 achieved by SGSN.

%The long-term prediction results are summarized for 4 action categories in Table~\ref{table:h36m_short_long}. Our method achieves state-of-the-art performance in the `discussion' category, and within the top two in all other categories. On average, it yields 58.75 and 81.95 
%for $r\!\!=\!\!560$ and 1000, respectively. 
%------------------------------------
\begin{table*}
\centering %\scriptsize
\footnotesize
\begin{tabular}{l|ccccc|ccccc}
\hline %\hline
actions & \multicolumn{5}{c}{basketball} & \multicolumn{5}{|c}{jumping} \\
\hline
milliseconds & 80 & 160 & 320 & 400 & 1000 & 80 & 160 & 320 & 400 & 1000 \\
\hline

%ResChunk-
1L & 12.81 & 21.28 & 39.10 & 51.74 & 98.61 & 15.63 & 30.55 & 56.92 & 71.05 & 142.33 \\

%ResChunk-
Fixed & 11.23 & 18.20 & 41.23 & 50.99 & 100.62 & 15.33 & 31.85 & 56.12 & 75.08 & 146.38 \\

%ResChunk-
1ch & 10.65 & 18.62 & 37.74 & 48.70 & 94.33 & 14.05 & 26.12 & 55.95 & 70.21 & 136.84 \\

%ResChunk-
4ch & 11.14 & 18.52 & 38.91 & 47.71 & 93.83 & 13.87 & 25.64 & \underline{54.19} & 68.29 & 131.06  \\

%ResChunk-
7ch & \textbf{09.83} & \textbf{17.20} & \underline{37.54} & \underline{47.39} & 95.20 & \textbf{12.27} & \underline{25.32} & 54.61 & \underline{68.12} & 132.45  \\

%ResChunk-
NoPONO & 10.89 & 19.26 & 38.78 & 48.76 & \underline{92.30} & 13.98 & 26.59 & 55.48 & 68.85 & \underline{130.91}  \\

ResChunk &
\underline{10.18} & \underline{17.42} & \textbf{37.33} & \textbf{45.28} & \textbf{86.88} & \underline{13.64} & \textbf{25.07} & \textbf{52.03} & \textbf{63.37} & \textbf{121.38} \\

\hline 
\end{tabular}
\caption{Influence of 
% different 
aspects of the proposed method in terms of MPJPE %(smaller is better) 
on two representative actions of the CMU Mocap dataset. %ResChunk-
1L, %ResChunk-
Fixed, %ResChunk-
1ch, %Reschunk-
4ch, %ResChunk-
7ch, %ResChunk- 
NoPONO, respectively correspond to single scale ResChunk, ResChunk with fixed and predefined groups, ResChunk with 1 output chunk, ResChunk with 4 output chunks, ResChunk with 7 output chunks, and ResChunk without position normalization. }
\label{table:ablation}
\end{table*}

\subsubsection{Discussion}

The quantitative comparisons show that our method consistently 
improves %both short-term and long-term 
motion prediction results. In CMU Mocap, for instance, we obtain the best performance in $67.5\%$ of cases, and best or second best in $100\%$ of cases,
while in H3.6M
we performed best in $80\%$
and best or second best in $100\%$ of cases.

One major aspect that sets our technique apart from others is the autoregressive modeling of the output variables via residual connections between subsequent output chunks. This, together with the ability to learn rich multiscale spatio-temporal motion features, allows our method to predict future sequences with greater accuracy.

Overall, two GCN-based methods of MSR-GCN and SGSN work better than the other three comparison methods. MSR-GCN uses a multiscale architecture to extract representative features in both fine-to-coarse and coarse-to-fine scales and imposes intermediate supervision at each scale. SGSN marginally outperforms MSR-GSN in most categories. It uses an adaptive graph scattering block to decompose information into various 
%graph spectrum 
bands, update the trainable features, and aggregate them via trainable attention weights. 

As we encode spatio-temporal motion features using both residual connections and dynamic adoptive multiscale GCN architecture, other methods encode temporal information via RNN~\cite{martinez2017human}, temporal convolutions~\cite{li2020dynamic},  DCT~\cite{mao2019learning,li2021skeleton}, or multiscale GCN architecture~\cite{dang2021msr}.
The temporal context is calculated frame-by-frame using RNN, resulting in the gradient vanishing or exploding. Temporal convolutions depend on the manually-defined size of the filters. The MSR-GCN architecture is similar to TrajDep except for the use of DCT for temporal encoding. It demonstrates that using a multiscale GCN architecture and computing global residuals between padded input poses and target poses directly, without converting to DCT coefficients, is effective and computationally efficient. 

The MSR-GCN 
however 
predefines the joint groups for the coarser scales which is not flexible to model dynamic sequences with different movement patterns. In our ablation experiment,
%in the next subsection, 
we study and 
quantify the impact of dynamic modeling 
%the movement patterns 
instead of fixed predefined groups. 

Moreover, our residual connections allow us to model the autoregressive temporal dependencies without having to explicitly define the range of temporal dependencies (or the temporal convolutional filter size, as done in prior work).

%-----------------------------------------------------
\subsection{Ablation Studies}

% To evaluate the contribution of different aspects of the proposed method, 
We conduct ablation studies to isolate aspects of the method, on both short-term and long-term motion predictions on the CMU Mocap dataset. 
In particular, we study the impact of (1) multiscale prediction, (2) automatic and 
adaptive joint grouping (3) residual chunks, (4) the number of chunks, and (5) position normalization (PONO).

\subsubsection{Multiple Scales}
We verify the impact of intermediate supervision by the coarser scale by simply removing the corresponding network and losses. We call this network \emph{1L} which in particular has only one path from $x_0$ to generate $y_0$. As a result, we observe a performance drop as shown in the first row of the Table~\ref{table:ablation}. 
Our method hence performs better than \emph{1L} by one scale level, proving its validity. 

\subsubsection{Automatic Joint Grouping}
To show the effectiveness of our grouping strategy, we choose a set of fixed groups based on~\cite{dang2021msr} and remove the corresponding joint grouping network and its related KL loss. We however select the same path to predict $y_1$. We call this network as \emph{Fixed}. As shown in the second row of the Table~\ref{table:ablation}, we notice that fixed grouping hurts the performance, although it outperforms the first experiment where we totally remove the multiscale prediction.

\subsubsection{Residual Chunks}
We remove the residual connections between sequence chunks, and replace them with one global residual between input and output sequences. The results are reported in the third row of the table where the new network, \emph{1ch}, marginally outperforms the \emph{Fixed}. 
The results demonstrate the benefit of %the 
residual connections specially for the long-term prediction.

\subsubsection{Number of Chunks}
In a separate experiment, we modify the number of residual connections and hence the number of sequence chunks. We 
%specifically 
create two networks called \emph{4ch} and \emph{7ch}, denoting ResChunk with 4 chunks and 7 chunks, respectively. As shown in Table~\ref{table:ablation}, increasing the number of chunks enhances the performance for the short-term prediction. It however negatively impacts the long-term prediction as the performance drops when we integrate 7 chunks.

\subsubsection{PONO}
We assess the impact of PONO by replacing the concatenation and PONO modules with a sum module.
In particular, $cat_i$ and $P_i$ 
%in Figure~\ref{fig:model} 
are removed and the input and output of $G_i$ are summed. % We call this new network as $NoPONO$ 
the results are shown in the 6-th row of the table (\emph{NoPONO}). We discovered that, while PONO has a minor influence, it is beneficial for more accurate outcomes.

%-----------------------------------------------------
%\subsection{Qualitative Results}

%To visually evaluate our proposed method, we illustrate qualitative comparisons with the state-of-the-art methods, as shown in Figure~\ref{fig:pred_results}. 
%Since the Zero-velocity~\cite{martinez2017human} baseline constantly predicts the last observed frame, it is unable to accurately predict complicated sequences even in short-term prediction. 
%Although Seq2seq~\cite{martinez2017human} and RNN-SPL~\cite{aksan2019structured} show relatively accurate short-term predictions, they suffer from discontinuities in longer predictions. 
%ResChunk, however, can generate more precise motions than the other methods since it maintains the temporal consistency of the motion structure in addition to the joint spatial relationships. %We further show the qualitative analysis of our method on two challenging sequences. 
%As shown, predictions are close to the ground-truths specially in short-term prediction.  

%\begin{figure}[!t]
%    \centering
%    \includegraphics[width=0.9\linewidth]{motion_result.pdf}
%    \caption{Qualitative results on CMU Mocap dataset.}
%    \label{fig:pred_results}
%\end{figure}

%-----------------------------------------------------
\section{Conclusion}
\label{sec:conclusion}
We propose ResChunk as a novel motion prediction method. It is trained to learn residuals between consecutive target sequence chunks as well as a dynamic action-agnostic multiscale representation for individual sequences. We perform rigorous experiments on two widely used datasets, CMU Mocap and H3.6M and compare 
% the performance of ResChunk 
its performance
to other leading related methods.
%in the area. 
Experiments show that our model outperforms other solutions to set a new state-of-the-art. 

We believe that the ability for our method to simultaneously learn temporal and spatial relationships provides opportunities to further explore this approach for other similar applications including time series forecasting and future event prediction in video sequences.

\bibliography{ref}

\clearpage
\appendix{APPENDIX}

In Sec.~\ref{sec:related_work}, we first introduce more related works.
We provide qualitative results of the proposed method in Sec.~\ref{sec:qualitative}. In Sec.\ref{sec:results}, we then show the prediction results on all 15 activities of the Human3.6M (H3.6M)~\cite{ionescu2013human3} dataset. 

\begin{table*}
\centering \scriptsize
\setlength{\tabcolsep}{2pt}
\begin{tabular}{l|cccc|cccc|cccc|cccc}
\hline
 & \multicolumn{4}{|c|}{walking} & \multicolumn{4}{|c|}{eating} & \multicolumn{4}{|c|}{smoking} & \multicolumn{4}{|c}{discussion}
 \\
\hline
milliseconds & 80 & 160 & 320 & 400 & 80 & 160 & 320 & 400 & 80 & 160 & 320 & 400 & 80 & 160 & 320 & 400 \\
\hline

Seq2seq~\cite{martinez2017human}
& 29.36 & 50.82 & 76.03 & 81.51 & 16.84 & 30.60 & 56.92 & 68.65 & 22.96 & 42.64 & 70.14 & 82.68 &  32.94 & 61.18 & 90.92 & 96.19  \\

TrajDep~\cite{mao2019learning}
& 12.29 & 23.03 & 39.77 & 46.12 & 08.36 & \underline{16.90} & \underline{33.19} & 40.70 & 07.94 & 16.24 & 31.90 & 38.90 & 12.50 & 27.40 & 58.51 & 71.68  \\

DMGNN~\cite{li2020dynamic} 
& 17.32 & 30.67 & 54.56 & 65.20 & 10.96 & 21.39 & 36.18 & 43.88 & 08.97 & 17.62 & 32.05 & 40.30 & 17.33 & 34.78 & 61.03 & 69.80 \\

MSR-GCN~\cite{dang2021msr}
& 12.16 & 22.65 & 38.64 & 45.24 & 08.39 & 17.05 & \textbf{33.03} & \underline{40.43} & 08.02 & 16.27 & 31.32 & 38.15 & 11.98 & 26.76 & 57.08 & 69.74  \\

SGSN~\cite{li2021skeleton}
& \underline{08.30} & \underline{15.00} & \textbf{26.70} & \textbf{31.50} & \underline{07.90} & 17.40 & 35.80 & 43.70 & \textbf{07.00} & \underline{13.80} & \textbf{23.60} & \textbf{28.20} & \underline{08.00} & \underline{19.10} & \textbf{34.70} & \textbf{40.40} \\

Ours (ResChunk) 
& \textbf{08.25} & \textbf{14.52} & \underline{26.91} & \underline{40.17} & \textbf{07.62} & \textbf{15.38} & {33.50} & \textbf{40.08} & \underline{07.85} & \textbf{13.41} & \underline{26.65} & \underline{33.94} & \textbf{07.20} & \textbf{18.25} & \underline{37.82} & \underline{45.16} \\ 

\hline \hline
& \multicolumn{4}{|c|}{directions} & \multicolumn{4}{|c|}{greeting} & \multicolumn{4}{|c|}{phoning} & \multicolumn{4}{|c}{posing} \\ \hline

milliseconds & 80 & 160 & 320 & 400 & 80 & 160 & 320 & 400 & 80 & 160 & 320 & 400 & 80 & 160 & 320 & 400 \\
\hline

Seq2seq~\cite{martinez2017human} &
35.36 & 57.27 & 76.30 & 87.67 & 34.46 & 63.36 & 124.60 & 142.50 & 37.96 & 69.32 & 115.00 & 126.73 & 36.10 & 69.12 & 130.46 & 157.08 \\

TrajDep~\cite{mao2019learning}
& \underline{08.97} & 19.87 & 43.35 & \underline{53.74} & 18.65 & 38.68 & 77.74 & 93.39 & 10.24 & 21.02 & 42.54 & 52.30 & 13.66 & 29.89 & 66.62 & 84.05 \\

DMGNN~\cite{li2020dynamic} 
& 13.14 & 24.62 & 64.68 & 81.86 & 23.30 & 50.32 & 107.30 & 132.10 & 12.47 & 25.77 & 48.08 & 58.29 & 15.27 & 29.27 & 71.54 & 96.65 \\

MSR-GCN~\cite{dang2021msr}
& \textbf{08.61} & \textbf{19.65} & \underline{43.28} & 53.82 & 16.48 & 36.95 & 77.32 & 93.38 & \underline{10.10} & 20.74 & 41.51 & 51.26 & 12.79 & 29.38 & 66.95 & 85.01 \\

SGSN~\cite{li2021skeleton}
& 10.60 & 21.60 & 45.40 & 56.30 & \underline{12.60} & \underline{26.50} & \underline{63.80} & \underline{79.60} & 10.90 & \textbf{18.10} & \underline{36.20} & \underline{41.40} & \textbf{08.20} & \underline{22.70} & \underline{64.80} & \underline{80.90} \\

Ours (ResChunk) 
& {09.32} & \underline{19.71} & \textbf{42.86} & \textbf{51.76} & \textbf{11.57} & \textbf{25.70} & \textbf{61.92} & \textbf{77.94} & \textbf{09.45} & \underline{19.76} & \textbf{34.88} & \textbf{40.61} & \underline{10.67} & \textbf{20.15} & \textbf{61.69} & \textbf{78.46} \\ 

\hline\hline
& \multicolumn{4}{|c|}{purchases} & \multicolumn{4}{|c|}{sitting} & \multicolumn{4}{|c|}{sittingdown} & \multicolumn{4}{|c}{takingphoto} \\ \hline

milliseconds & 80 & 160 & 320 & 400 & 80 & 160 & 320 & 400 & 80 & 160 & 320 & 400 & 80 & 160 & 320 & 400 \\
\hline

Seq2seq~\cite{martinez2017human} &
36.33 & 60.30 & 86.53 & 95.92 & 42.55 & 81.40 & 134.70 & 151.78 & 47.28 & 85.95 & 145.75 & 168.86 & 26.10 & 47.61 & 81.40 & 94.73 \\

TrajDep~\cite{mao2019learning}
& 15.60 & 32.78 & \underline{65.72} & 79.25 & 10.62 & \underline{21.90} & 46.33 & 57.91 & 16.14 & 31.12 & 61.47 & 75.46 & 09.88 & 20.89 & 44.95 & 56.58 \\

DMGNN~\cite{li2020dynamic} 
& 21.35 & 38.71 & 75.67 & 92.74 & 11.92 & 25.11 & \textbf{44.59} & \textbf{50.20} & 14.95 & 32.88 & 77.06 & 93.00 & 13.61 & 28.95 & 45.99 & 58.76 \\

MSR-GCN~\cite{dang2021msr}
& \underline{14.75} & \textbf{32.39} & 66.13 & 79.64 & 10.53 & 21.99 & 46.26 & 57.80 & 16.10 & 31.63 & 62.45 & 76.84 & 09.89 & 21.01 & 44.56 & 56.30 \\

SGSN~\cite{li2021skeleton}
& 18.40 & 36.90 & \textbf{60.00} & \textbf{68.50} & \underline{09.80} & 23.00 & 46.20 & 56.40 & \underline{10.10} & \textbf{24.70} & \underline{51.00} & \underline{60.20} & \textbf{06.00} & \textbf{13.90} & \underline{36.30} & \textbf{47.80} \\

Ours (ResChunk) 
& \textbf{13.37} & \underline{32.70} & 65.63 & \underline{74.56} & \textbf{09.34} & \textbf{20.73} & \underline{45.84} & \underline{53.97} & \textbf{09.18} & \underline{27.86} & \textbf{48.16} & \textbf{59.90} & \underline{07.49} & \underline{15.61} & \textbf{35.83} & \underline{51.28} \\

\hline \hline
& \multicolumn{4}{|c|}{waiting} & \multicolumn{4}{|c|}{walkingdog} & \multicolumn{4}{|c|}{walkingtogether} & \multicolumn{4}{|c}{Average} \\ \hline

milliseconds & 80 & 160 & 320 & 400 & 80 & 160 & 320 & 400 & 80 & 160 & 320 & 400 & 80 & 160 & 320 & 400 \\
\hline 

Seq2seq~\cite{martinez2017human} &
30.62 & 57.82 & 106.22 & 121.45 & 64.18 & 102.10 & 141.07 & 164.35 & 26.79 & 50.07 & 80.16 & 92.23 & 34.66 & 61.97 & 101.08 & 115.49 \\

TrajDep~\cite{mao2019learning}
& 11.43 & 23.99 & 50.06 & 61.48 & 23.39 & 46.17 & 83.47 & 95.96 & 10.47 & 21.04 & 38.47 & 45.19 & 12.68 & 26.06 & 52.27 & 63.51 \\

DMGNN~\cite{li2020dynamic} 
& 12.20 & 24.17 & 59.62 & 77.54 & 47.09 & 93.33 & 160.13 & 171.20 & 14.34 & 26.67 & 50.08 & 63.22 & 16.95 & 33.62 & 65.90 & 79.65 \\

MSR-GCN~\cite{dang2021msr}
& 10.68 & 23.06 & \underline{48.25} & \underline{59.23} & \underline{20.65} & \underline{42.88} & \textbf{80.35} & \textbf{93.31} & 10.56 & 20.92 & 37.40 & 43.85 & 12.11 & 25.56 & 51.64 & 62.93 \\

SGSN~\cite{li2021skeleton}
& \underline{08.10} & \underline{20.10} & 52.80 & 67.80 & 25.70 & 53.00 & 93.00 & 111.40 & \textbf{08.10} & \underline{17.60} & \textbf{34.50} & \underline{43.80} & \underline{10.65} & \underline{22.90} & \underline{47.00} & \underline{56.90} \\

Ours (ResChunk) 
& \textbf{07.47} & \textbf{19.87} & \textbf{45.64} & \textbf{59.66} & \textbf{18.53} & \textbf{40.81} & \underline{81.25} & \underline{95.80} & \underline{08.96} & \textbf{16.55} & \underline{36.76} & \textbf{42.93} & \textbf{09.15} & \textbf{21.40} & \textbf{45.69} & \textbf{56.41} \\ \hline

\end{tabular}
\caption{Short-term prediction results over 15 action categories of H3.6M dataset. The average prediction results across all the actions are also shown. The results are reported in terms of MPJPE (smaller is better). The best performance is in boldface, and the second-best performance is underlined.}
\label{table:h36m_short}
\end{table*}

\section{Related Work}
\label{sec:related_work}
We include additional related work that we were unable to cover in the main paper due to space constraints. Some of the other most relevant techniques to our work, in addition to the works already mentioned (such as MSR-GCN~\cite{dang2021msr} and Cui~\etal~\cite{cui2020learning}), can be listed as seq2seq model~\cite{martinez2017human}, DMGNN~\cite{li2020dynamic}, 
%MSR-GCN~\cite{dang2021msr}, Cui~\etal~\cite{cui2020learning}, 
and GAGCN~\cite{zhong2022spatio}. 
%have previously been addressed, thus the remainder are provided here. 

Martinez~\etal~\cite{martinez2017human}  developed a sequence-to-sequence architecture with residual connections by applying changes such as sample-based loss to the standard RNN. Comprehensive motion modeling, however, requires both spatial and temporal reasoning, while RNN-based methods are generally unable to model the spatial relationships between the joints. Moreover, the training of an RNN model easily collapses via gradient explosion. Discontinuities are also common due to the frame-wise prediction style.

DMGNN~\cite{li2020dynamic} modeled the internal relationships among body joints. 
It characterized the human body at multiple scales by grouping spatially proximal joints together and representing each group by a coarser pose. It was further improved in MST-GNN~\cite{li2021multiscale} by adding a graph-based attention GRU and trainable spatial graph pooling. 
They are however different from our method that automatically groups together even distant joints not directly connected by the kinematic tree if they similarly participate in a particular movement.

GAGCN~\cite{zhong2022spatio} learns the complex spatiotemporal dependencies over diverse action types. 
While our method focuses on extracting correlated body components at multiple scales in individual sequences, GAGCN addresses the issue of cross-dependency of space and time by balancing the weights of spatio-temporal modeling via scaling the number of candidate matrices. 

Our automatic scaling and the concept of body components are also different from 
Cui~\etal~\cite{cui2020learning}
and DMGNN~\cite{li2020dynamic} 
as distant joints can be grouped together in our method, and produced components may differ in two individual sequences. 

%As opposed to this method, we dynamically learn the relationships between all pairs of joints to extract correlated body components at multiple scales in individual sequences. Specifically, body components are not fixed and might be different in two sequences.

Significantly and distinct from all other methods, our method is trained to learn the residuals between target sequence chunks in an autoregressive manner to enforce the temporal connectivities between consecutive chunks. 
%We specifically learn the \emph{residual between target sequence chunks autoregressively}. 
Other methods such as~\cite{cui2020learning,mao2019learning,dang2021msr}, 
also use residual connections by adding global skip connections between the input and the output poses of their networks. We however learn the residual vectors between the corresponding sequence chunks in an efficient autoregressive fashion in a single forward pass. Specifically, we benefit from both autoregressive learning and residual connections.

%These methods try to dynamically learn the joint connections. As explained in response to the previous comment, our method however is fundamentally different as we \emph{dynamically} learn the relationships between all pairs of joints to extract \emph{correlated body components} at \emph{multiple scales} in \emph{individual sequences}. 

%-----------------------------------------------------
\section{Qualitative Results}
\label{sec:qualitative}
\begin{figure*}%[H]
    \centering
    \includegraphics[width=0.8\linewidth]{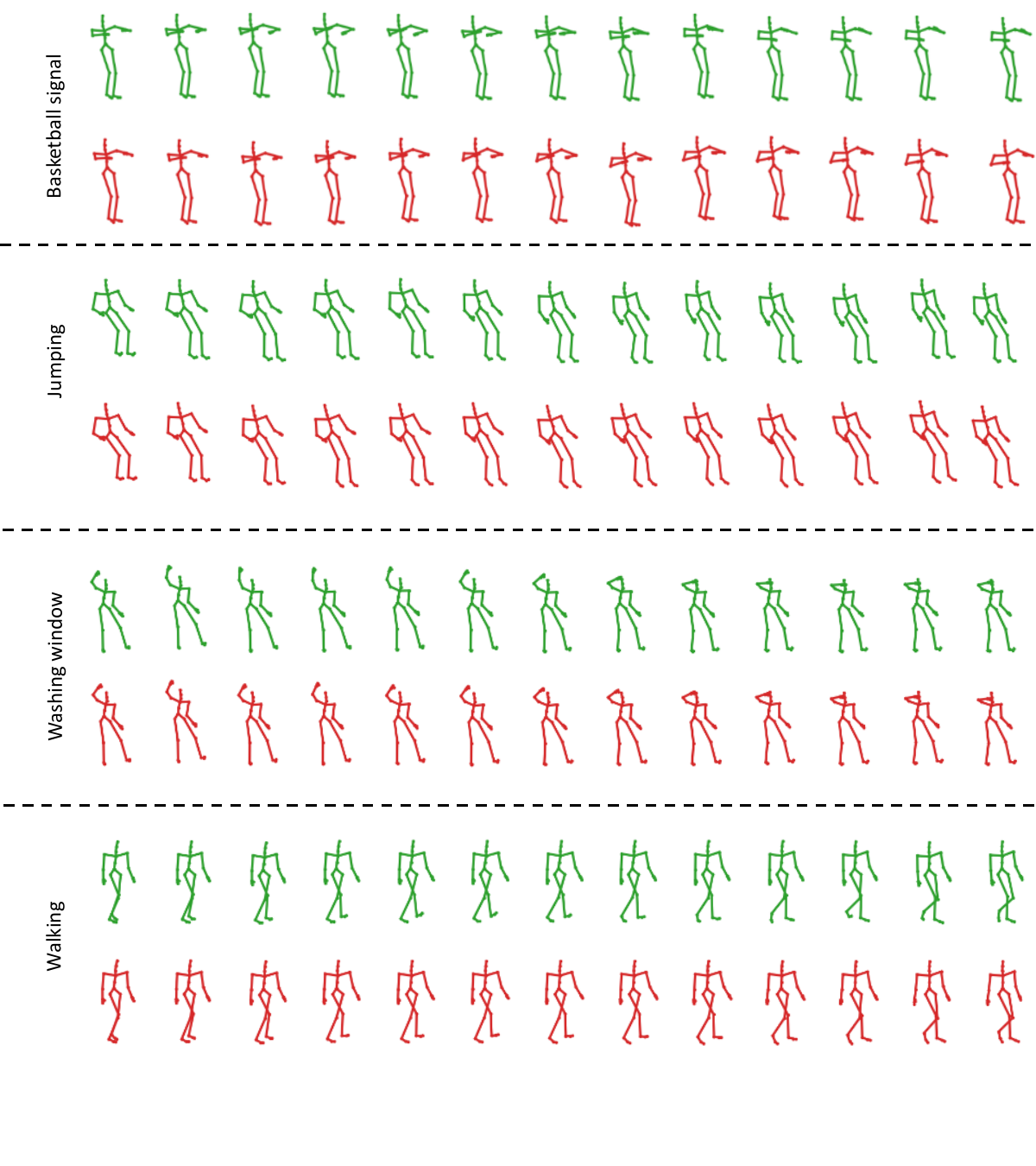}
    \caption{Qualitative results over 1000 ms are shown for basketball signal, jumping, washing window, and walking on the CMU Mocap dataset. Ground-truths and predictions results are displayed in green and red colors, respectively.}
    \label{fig:cmu_results}
\end{figure*}

\begin{figure*}%[H]
    \centering
    \includegraphics[width=0.8\linewidth]{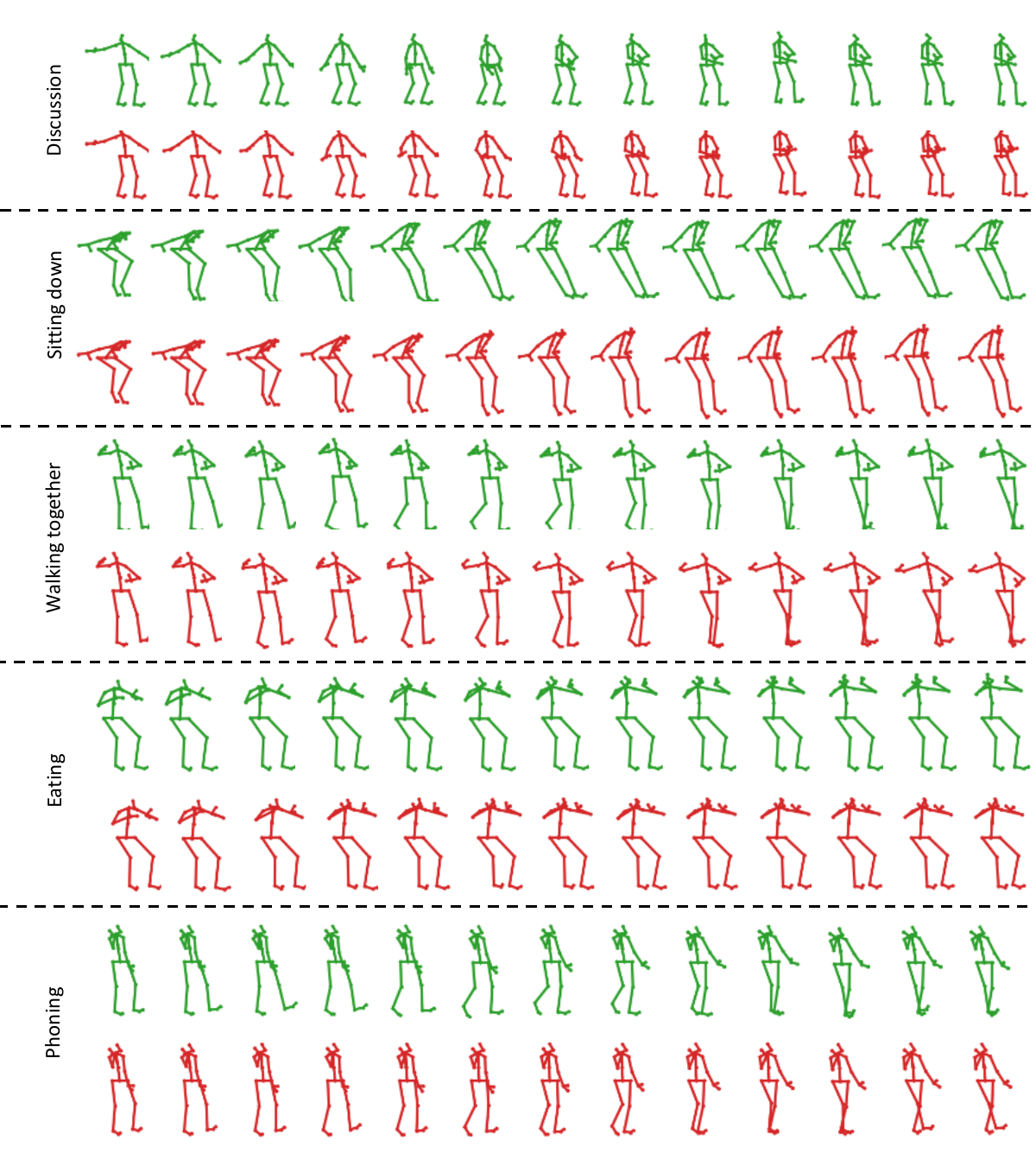}
    \caption{Qualitative results over 1000 ms are shown for discussion, sitting down, walking together, eating, and phoning on the H3.6M dataset. Ground-truths and predictions results are displayed in green and red colors, respectively.}
    \label{fig:h36m_results}
\end{figure*}

To visually evaluate our proposed method, we illustrate some representative predictions frame by frame over 1000 ms in the future. Figure~\ref{fig:cmu_results} visualizes the motion prediction results on the CMU motion capture (CMU Mocap, http://mocap.cs.cmu.edu/) dataset. The predictions are shown for 4 activities of \emph{basketball signal}, \emph{jumping}, \emph{washing window}, and \emph{walking}. 
Additionally, we show some representative results on 5 activities of the H3.6M dataset in Figure~\ref{fig:h36m_results}. The visualized actions include \emph{discussion}, \emph{sitting down}, \emph{walking together}, \emph{eating}, and \emph{phoning}. 
As shown in both figures, predictions are close to the ground-truths specially in short-term prediction. The results show that our method can maintain the temporal consistency of the motion structure in addition to the joint spatial relationships.

\begin{table*}%[h!]
\centering \scriptsize
\setlength{\tabcolsep}{2pt}
\begin{tabular}{l|cc|cc|cc|cc|cc}
\hline
 & \multicolumn{2}{|c}{eating} & \multicolumn{2}{|c}{smoking} & \multicolumn{2}{|c}{discussion} & \multicolumn{2}{|c}{directions} &
 \multicolumn{2}{|c}{Average}
 \\
\hline
milliseconds & 560 & 1000 & 560 & 1000 & 560 & 1000 & 560 & 1000 & 560 & 1000  \\
\hline

Seq2seq~\cite{martinez2017human}
& 79.87 & 100.20 & 94.83 & 137.44 & 121.30 & 161.70 & 110.05 & 152.48 & 101.51 & 137.96  \\

TrajDep~\cite{mao2019learning}
& 53.39 & 77.75 & 50.74 & 72.62 & 91.61 & 121.53 & \textbf{71.01} & 101.79 & 66.69 & 93.42 \\

DMGNN~\cite{li2020dynamic} 
& 58.11 & 86.66 & 50.85 & 72.15 & 81.90 & 138.32 & 110.06 & 115.75 & 75.23 & 103.22 \\

MSR-GCN~\cite{dang2021msr}
& \underline{52.54} & 77.11 & 49.45 & 71.64 & 88.59 & 117.59 & \underline{71.18} & \underline{100.59} & 65.44 & 91.73  \\

SGSN~\cite{li2021skeleton} 
& 56.00 & \textbf{68.20} & \textbf{31.60} & \textbf{58.80} & \underline{69.10} & \underline{96.50} & 79.00 & 101.00 & \underline{58.93} & \textbf{81.13} \\

Ours (ResChunk) 
& \textbf{51.78} & \underline{70.54} & \underline{42.93} & \underline{63.25} & \textbf{68.41} & \textbf{95.36} & {71.87} & \textbf{98.65} & \textbf{58.75} & \underline{81.95} \\

\hline
\end{tabular}
\caption{Long-term prediction results over four representative action categories of H3.6M dataset. The average prediction results across all the actions are also shown. The results are reported in terms of MPJPE (smaller is better). The best performance is in boldface, and the second-best performance is underlined.}
\label{table:h36m_long}
\end{table*}

\section{Results on Human3.6M}
\label{sec:results}

Table~\ref{table:h36m_short} reports the short-term prediction results on the H3.6M dataset. As shown, our method achieves the best performance in most time horizons for all activities. On the `greeting' category, for example, ResChunk yields smaller MPJPE errors of 11.57, 25.70, 61.92, 77.94 respectively
for $r=80, 160, 320$, and $400$, which outperforms SGSN~\cite{li2021skeleton} with 12.60, 26.50, 63.80, 79.60 respectively in the same time horizons. On average, ResChunk improves the short-term prediction significantly by obtaining the best performance 
for all $r=80, 160, 320$, and $400$ horizons as 9.15, 21.40, 45.69, 56.41, respectively. 

The long-term prediction results are summarized for 4 action categories in Table~\ref{table:h36m_long}. Our method achieves state-of-the-art performance in the `discussion' category, and the best or the second-best performance in all other categories. On average, it yields 58.75 and 81.95 
for $r=560$ and 1000, respectively.

\end{document}